\documentclass[letterpaper,10pt,journal,twoside]{ieeetran} 
\IEEEoverridecommandlockouts                              % This command is only needed if 
\pdfoutput=1

\usepackage{bm}
\usepackage{amsmath,amssymb,amsthm}
\usepackage[table]{xcolor}
\usepackage[caption=false]{subfig}
\usepackage[]{mathtools}
\usepackage{bbm}
\usepackage{mdframed}
\usepackage[sort,compress]{cite} % Use [noadjust] instead?
\usepackage{tcolorbox}
\usepackage{algorithmicx}
\usepackage{algorithm}
\usepackage[noend]{algpseudocode}
\usepackage{setspace}

\newcommand*\Let[2]{\State #1 $\gets$ #2}
\algrenewcommand\algorithmicrequire{\textbf{Input:}}
\algrenewcommand\algorithmicensure{\textbf{Output:}}
\algnewcommand\algorithmicinput{\textbf{Input:}}
\algnewcommand\INPUT{\item[\algorithmicinput]}
\usepackage{moreverb,url}
\usepackage[colorlinks]{hyperref}
\usepackage{icomma} 
\usepackage{pgfplots}
\usepgfplotslibrary{fillbetween}
\usepackage[perpage]{footmisc}
\usepackage{soul}
\usepackage{fontawesome}
\DeclareFontFamily{OT1}{pzc}{}
\DeclareFontShape{OT1}{pzc}{m}{it}{<-> s * [1.200] pzcmi7t}{}
\DeclareMathAlphabet{\mathpzc}{OT1}{pzc}{m}{it}

\AtBeginDocument{%
  \DeclareMathAlphabet\PazoBB{U}{fplmbb}{m}{n}%
}

\usepackage{dsfont}
\newtheorem{theorem}{Theorem}

\algrenewcommand\textproc{}%


\let\oldthempfootnote\thempfootnote
\def\thempfootnote{\text{\oldthempfootnote}}
\usepackage{enumitem}

\setlist[itemize]{leftmargin=*}
\setlist[enumerate]{leftmargin=*}

\colorlet{lightgray}{green!4}

\usepackage[keeplastbox]{flushend}
\usepackage{color}
\usepackage{xparse}

\newcommand{\norm}[2]{\|#1\|_{#2}}
\newcommand{\frob}[1]{\|#1\|_{\scriptscriptstyle F}}
\newcommand{\mbf}[1]{\mathbf{#1}}

\NewDocumentCommand\bbm{}{ \begin{bmatrix} }
\NewDocumentCommand\ebm{}{ \end{bmatrix} }

\NewDocumentCommand\Matrix{m}{ \boldsymbol{\mathbf{#1}} }

\NewDocumentCommand\CoordinateFrame{m}{ \underrightarrow{\Matrix{\mathcal{F}}}_{#1} }

\include{matt_giamou_notation}

\newcommand{\jk}[1]{}
\newcommand{\vp}[1]{}

\usepackage{soul}
\usepackage[outdir=./]{epstopdf}
\usepackage[para]{threeparttable}
\usepackage{multirow}
\usepackage{booktabs}

\title{\Large \bf
Certifiably Globally Optimal Extrinsic Calibration \\from Per-Sensor Egomotion \jk{title bigger for RA-L, no?}
}

\author{Matthew Giamou$^{1}$, Ziye Ma$^{1}$, Valentin Peretroukhin$^{1}$, and Jonathan Kelly$^{1}$ \thanks{Manuscript received: September 9, 2018; Revised November 19, 2018; Accepted December 10, 2018.}%Use only for final RAL version
\thanks{This paper was recommended for publication by Editor Dezhen Song  upon evaluation of the Associate Editor and Reviewers' comments. This work was supported in part by the Natural Sciences and Engineering Research Council of Canada and by a Dean's Catalyst Professorship from the University of Toronto.}%Use only for final RAL version
\thanks{$^{1}$All authors are with the Space \& Terrestrial Autonomous Robotics Systems (STARS) laboratory at the University of Toronto Institute for Aerospace Studies (UTIAS), Canada. {\tt\small <firstname>.<lastname>@robotics.utias.utoronto.ca}}%
\thanks{Digital Object Identifier (DOI): see top of this page.} }
\begin{document}
\maketitle
%\thispagestyle{empty}
%\pagestyle{empty}

% Paper headers
\markboth{IEEE Robotics and Automation Letters. Preprint Version. Accepted December, 2019}
{Giamou \MakeLowercase{\textit{et al.}}: Certifiably Globally Optimal Extrinsic Calibration} % Use only for final RAL version

%%%%%%%%%%%%%%%%%%%%%%%%%%%%%%%%%%%%%%%%%%%%%%%%%%%%%%%%%%%%%%%%%%%%%%%%%%%%%%%%
\begin{abstract}
We present a certifiably globally optimal algorithm for determining the extrinsic calibration between two sensors that are capable of producing independent egomotion estimates. This problem has been previously solved using a variety of techniques, including local optimization approaches that have no formal global optimality guarantees. We use a quadratic objective function to formulate calibration as a quadratically constrained quadratic program (QCQP). By leveraging recent advances in the optimization of QCQPs, we are able to use existing semidefinite program (SDP) solvers to obtain a certifiably global optimum via the Lagrangian dual problem. Our problem formulation can be globally optimized by existing general-purpose solvers in less than a second, regardless of the number of measurements available and the noise level. This enables a variety of robotic platforms to rapidly and robustly compute and certify a globally optimal set of calibration parameters without a prior estimate or operator intervention. We compare the performance of our approach with a local solver on extensive simulations and multiple real datasets. Finally, we present necessary observability conditions that connect our approach to recent theoretical results and analytically support the empirical performance of our system.\jk{what's the sell in one sentence? faster, better, cheaper? why do I as a robo-person care? I think this can be inserted right after "and the noise level."  ... something like ``This enable <such and such>'' ... maybe use word `best'}
\end{abstract}

%%%%%%%%%%%%%%%%%%%%%%%%%%%%%%%%%%%%%%%%%%%%%%%%%%%%%%%%%%%%%%%%%%%%%%%%%%%%%%%%
% Keywords appear just beneath the abstract. Use only for final RAL version.  

% Do these have to come from the list on the submission? 
% Use these for RAL submission

\section{Introduction}
\IEEEPARstart{R}{obots} rely on calibrated sensors in order to safely and effectively carry out complex tasks. For mobile robots equipped with multiple instruments, an accurate estimate of the relative pose (extrinsic calibration) between each pair of sensors is crucial in enabling capabilities like reliable localization and mapping. Commercial robots may ship with a factory calibration performed by experts using precision equipment that is unavailable to the end user. During operation, intentional adjustments or unintended mechanical stresses may necessitate recalibration in the field. This need for recalibration outside of a factory or laboratory setting has led to a plethora of automatic calibration methods for a variety of sensor combinations\cite{pandey2015automatic, brookshire2013extrinsic, kelly2011visual,lambert2017entropy}. These methods operate with varying speed, accuracy, and assumptions about the robot's environment, and also differ in the level of sensor specificity and technician involvement required. Robots capable of true long-term autonomy must leverage accurate and fully automatic calibration procedures that work in their deployed environments. However, most existing approaches do not come with any guarantee that globally optimal calibration parameters will be found (given the available data); convergence to a local minimum may result in very poor calibration quality, compromising both the reliability and safety of the robot system.

% and therefore require a reasonable initial guess.
% If adjustments to sensor placement are needed or unintentional variations are experienced by a robot in the field, the extrinsic calibration will need to be recomputed.

\begin{figure}[t]
	\centering
	\includegraphics[width=\columnwidth]{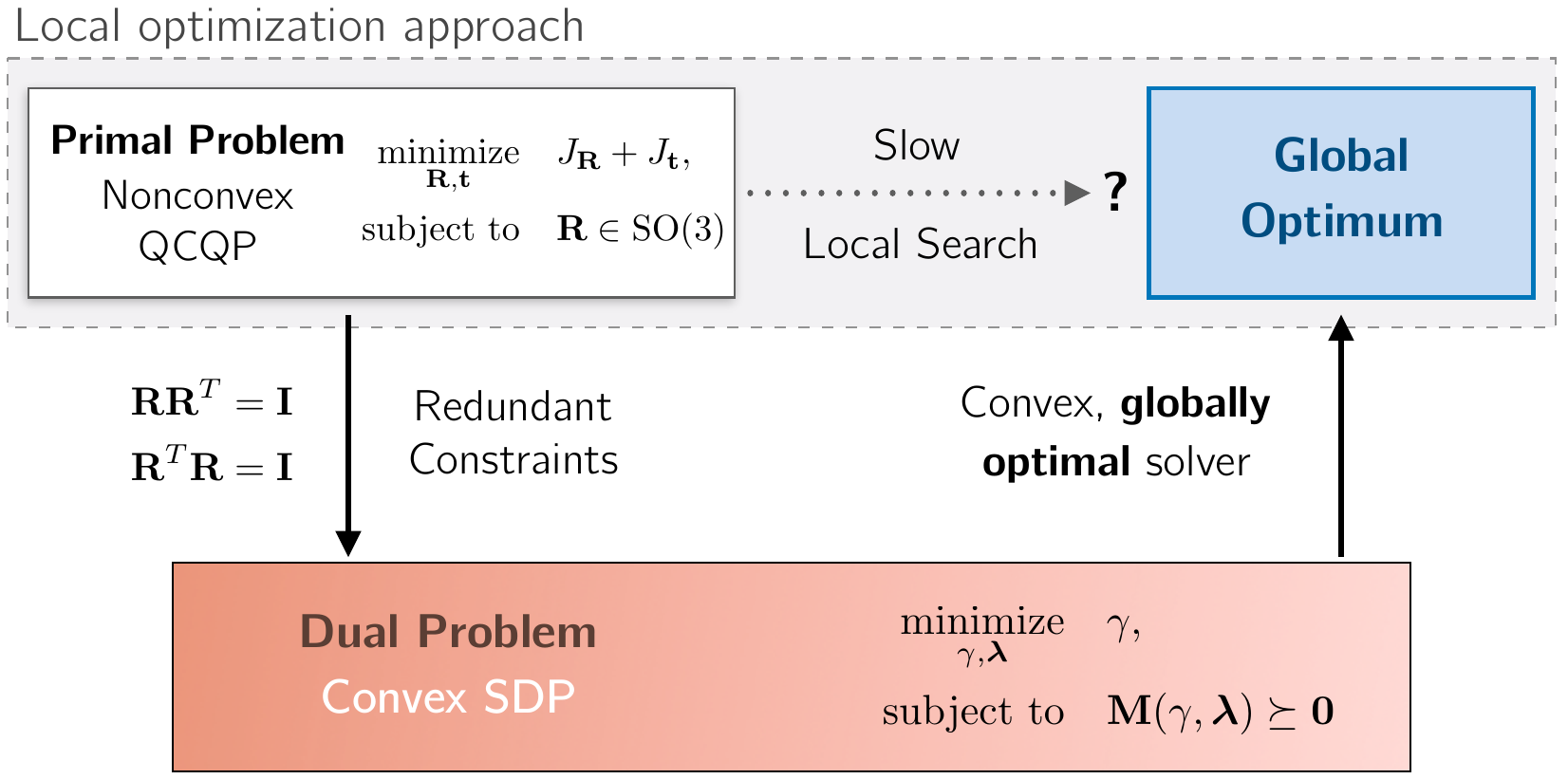}
	\caption{We compute the extrinsic calibration between two egomotion sensors by solving the Lagrangian dual of a nonconvex quadratically constrained quadratic program. The dual is a convex problem, meaning it can be efficiently solved to find the global optimum. We leverage recent theoretical results and prove that the primal solution can be extracted from the dual solution, even in the presence of very significant noise. This leads to a certifiably globally optimal approach that outperforms local optimization methods that have no formal guarantees.\jk{comment on diagram}}
	\label{fig:system}
	\vspace{0.2cm}
\end{figure}

In this paper, we develop a procedure for quickly determining the extrinsic calibration between any pair of sensors capable of producing egomotion estimates. Our problem formulation is closely related to \cite{brookshire2013extrinsic}, but we take a different approach by leveraging recent advances in the optimization of quadratically constrained quadratic programs (QCQPs). This enables us to provide a \emph{certifiably globally optimal solution} to our cost function, even when severe measurement noise is present, guaranteeing that our approach avoids local minima. The main contributions of our work are

\begin{enumerate}
\item a QCQP formulation of extrinsic calibration from per-sensor egomotion measurements,
\item a fast, certifiably globally optimal convex solution method for our formulation, 
\item a theorem connecting calibration observability and the tightness of our Lagrangian dual solution (see supplementary material for proofs), and 
\item an open source implementation and experimental analysis of our algorithm in MATLAB.\footnote[2]{See \url{http://github.com/utiasSTARS/certifiable-calibration} for code and supplementary material.}
\end{enumerate}

In our experiments we compare our work with a method that uses a maximum likelihood estimation (MLE) problem formulation and a local solver~\cite{brookshire2013extrinsic}. This comparison highlights the viability of our cost function in terms of representing the problem and getting an accurate estimate, while also demonstrating the speed and guaranteed avoidance of local minima that our formulation enables through global convex optimization. Additionally, the global optimality guarantees could be useful in providing a first estimate for a more precise method using more probabilistic information or dense reconstructions from sensor data, or to present a hypothesis that helps reject outliers for a robust estimation scheme. Our high-level goal is to demonstrate and experimentally verify desirable theoretical properties of our QCQP formulation of the extrinsic calibration problem, with an eye towards other estimation problems in robotics and computer vision. 

\section{Related Work}

A unified framework for calibration between sensors capable of providing egomotion estimates was developed for 2D and 3D motion respectively in \cite{brookshire2012automatic} and \cite{brookshire2013extrinsic}. In that framework, a MLE problem is formulated and solved with an iterative Levenberg-Marquardt solver. The closely related hand-eye calibration problem~\cite{horaud1995hand} has been extensively studied~\cite{condurache2016orthogonal, dornaika1998simultaneous}, and we use a cost function formulation similar to the one in \cite{heller2014hand}. 

\subsection{Globally Optimal Calibration}
The majority of automatic extrinsic calibration approaches do not guarantee that the global optimum will be found. In \cite{levinson2014unsupervised}, the extrinsic calibration of a multi-beam lidar sensor relative to a robot's base frame is computed with a local optimization method. The authors demonstrate that their approach avoids converging to inaccurate local optima, even when initialized with highly inaccurate parameter values, but they provide no formal guarantee that this condition holds in general. Certain bespoke algorithms involving specific sensor pairs and environmental features can be solved in closed form with a minimal set of measurements~\cite{gomez2015extrinsic,zhang2004extrinsic}, but these methods require a local optimization when noise is present. The approach for calibrating extrinsic sensor parameters relative to a manipulator base in \cite{limoyo2018self} relies on solving a point cloud registration problem. While there are globally optimal branch-and-bound (BnB) algorithms that can guarantee a solution to point cloud registration up to a desired accuracy~\cite{straub2017efficient}, these techniques can be extremely slow. Similarly, the global optimum of a hand-eye calibration problem is found in \cite{heller2012branch} and \cite{ruland2012globally} using BnB, but the runtime is orders of magnitude larger than for convex methods. 

The calibration methods closest to our own are found in \cite{heller2014hand} and \cite{heller2014world}, where certifiably globally optimal solutions for hand-eye calibration are recovered using the method of convex linear matrix inequality relaxations \cite{lasserre2001global}. However, in this paper we focus on mobile robotic platforms, certify global optimality for more severe noise, theoretically prove the global optimality of our approach with new convex optimization theory~\cite{cifuentes2017local}, and connect our approach with a well-known observability result. 

\subsection{Lagrangian Duality-Based Optimization}
\jk{I actually think these two paragraphs could be moved to problem formulation ... or they need a different sub-heading - they're background and not quite related work.} Recently, a number of certifiably correct solutions to estimation problems in robotics have been developed. In \cite{carlone2015lagrangian}, Lagrangian Duality was used for verifying whether a candidate solution to a pose graph optimization (PGO) is globally optimal. This approach led to fast solvers in \cite{rosen2016se} and \cite{briales2017cartan} that exploit the strong duality of PGO when measurement noise is not severe. The related problem of rotation averaging has also proven amenable to globally optimal duality-based solution methods in \cite{fredriksson2012simultaneous} and \cite{eriksson2018rotation}. %PGO and rotation averaging are key problems in computer vision and robotics that form the backbone of many systems.

Other problems involving optimization over SO(3) or SE(3) variables that can be solved via their Lagrangian dual include generalized point cloud registration with known correspondences~\cite{briales2017convex, olsson2008solving} and the relative camera pose problem~\cite{briales2018certifiably}. Both of these problems involve optimization over a single rotation argument, and both use the method of adding redundant orthogonality constraints found in \cite{anstreicher2000lagrangian} and \cite{wolkowicz2002note} for recovering a tight dual. In this work, we develop a problem structure similar to that of \cite{briales2017convex} and use the same basic solution procedure. 

\section{Problem Formulation}

\begin{figure}[t]
	\centering
	\includegraphics[width=\columnwidth]{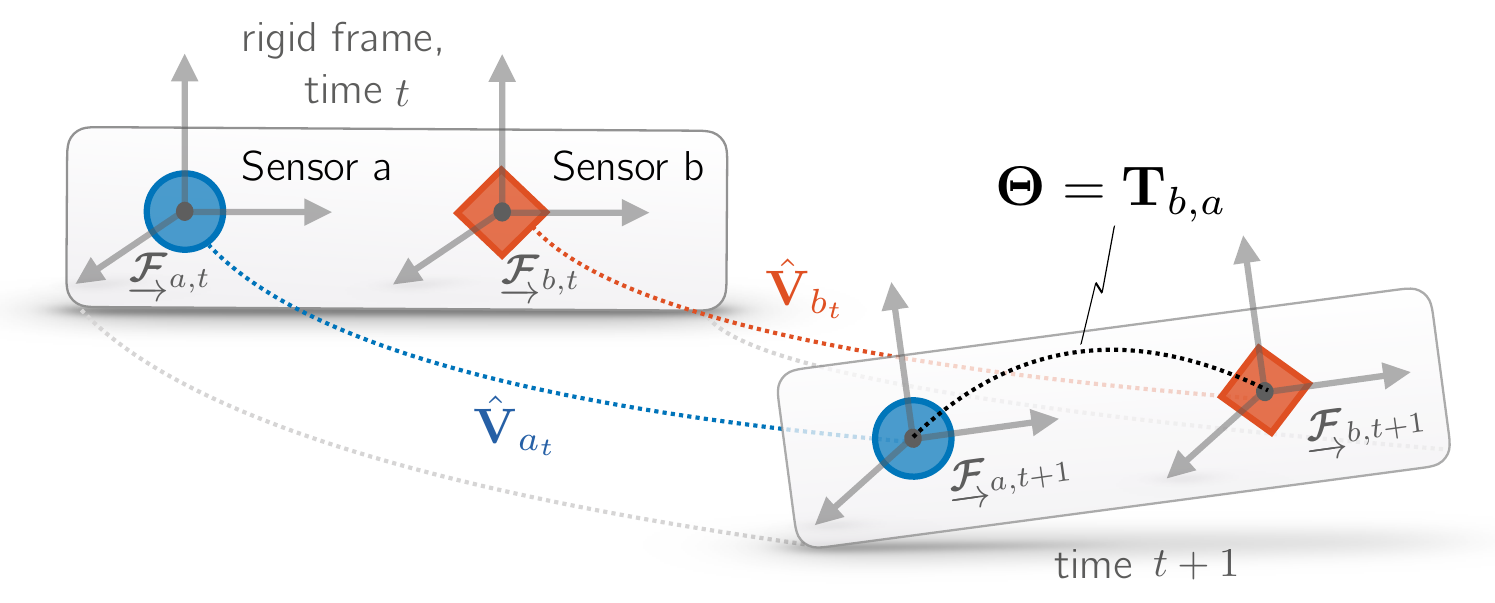}
	\caption{We perform extrinsic calibration by using two pairs of rigidly attached egomotion sensors. The extrinsic calibration $\bm{\Theta} \in \text{SE}(3)$ relates the poses of sensors $a$ and $b$, which are rigidly fixed to the same robot body during measurement acquisition. At each timestep $t$, the sensors produce independent egomotion estimates $\hat{\mbf{V}}_{a_t}, \hat{\mbf{V}}_{b_t} \in \text{SE}(3)$, which are used to estimate $\bm{\Theta}$. \jk{can make caption a bit better by defining what K = Tba is ... and V's etc. may as well use space we have.}}
	\label{fig:sensor_geometry}
	\vspace{0.2cm}
\end{figure}

In this section, we describe the notation, variables, and cost function used to formulate the optimization problem considered in the remainder of the paper. Figure \ref{fig:sensor_geometry} summarizes the kinematics used in our formulation.

\subsection{Notation}
Boldface is used for vector and matrix variables with lowercase and uppercase letters respectively (e.g. $\mbf{x}, \mbf{A}$). The vector representing the $i$th column of the matrix $\mbf{A}$ is denoted $\mbf{A}_i$. The $3\times3$ identity matrix is written $\mbf{I}$. The matrix of zeros of size $m\times n$ is denoted $\mbf{0}_{m\times n}$ (or without subscripts when the dimensions can be easily inferred from context). We use $\text{Sym}(d)$ to denote the set of $d\times d$ symmetric matrices. The homogeneous transformation matrix $\mbf{T}_{a,b}$ transforms the coordinates of a point expressed in reference frame $b$ into the coordinates of the same point expressed in frame $a$. We write $\mbf{A} \succeq \mbf{0}$ to indicate that $\mbf{A}$ is a positive semidefinite matrix and $\mbf{A} \succ \mbf{0}$ to indicate positive definiteness. The Kronecker product between two matrices is denoted $\mbf{A} \otimes \mbf{B}$. Finally,  $\mbf{A}/\mbf{B}$ denotes the Schur complement~\cite{zhang2006schur} of matrix $\mbf{A}$ with one of its diagonal sub-matrices $\mbf{B}$.

\subsection{Extrinsic Calibration from Per-Sensor Egomotion}

Our problem formulation is closely related to that in \cite{brookshire2013extrinsic}. Figure \ref{fig:sensor_geometry} displays the problem setup and relevant parameters. Sensors $a$ and $b$ are rigidly attached to a vehicle or rig. We denote the coordinate frames in which egomotion estimates are made at time step $t$ with $\CoordinateFrame{a,t}$ and $\CoordinateFrame{b,t}$. We assume that the sensors produce pairs of egomotion estimates represented by homogeneous transformation matrices $\hat{\mbf{T}}_{a_{t-1}, a_{t}}, \hat{\mbf{T}}_{b_{t-1}, b_{t}} \in \text{SE}(3)$ corresponding to the motion from time step $t-1$ to time step $t$, which can be achieved for unsynchronized sensors via an interpolation method~\cite{brookshire2012automatic}. 

We represent the constant extrinsic calibration between $\CoordinateFrame{a,t}$ and $\CoordinateFrame{b,t}$ with the homogeneous transformation matrix \jk{wanna add footnote why K?}
\begin{equation}
\bm{\Theta} = \mbf{T}_{b,a} =  \begin{bmatrix}
 \mbf{R} & \mbf{t}\\
\mbf{0} & 1
\end{bmatrix}.
\end{equation}
At any time $t$, the following relationship holds between $\bm{\Theta}$ and the absolute pose of sensors $a$ and $b$ in the world frame $\CoordinateFrame{w}$:
\begin{equation}
\mbf{T}_{a_{t}} = \mbf{T}_{w, a_{t}} = \mbf{T}_{b_{t}}\bm{\Theta} = \mbf{T}_{w, b_{t}}\bm{\Theta}.
\end{equation}
We simplify the notation for the true motion of sensor $s$ from time step $t-1$ to time step $t$ with 
\begin{align}
\mbf{T}_{s_{t-1}, s_{t}} &= \mbf{V}_{s_t}, \ s \in \{a, b\}, \\
&= \begin{bmatrix}
 	 \mbf{R}_{s_t} & \mbf{t}_{s_t} \\
 	 \mbf{0} & 1 
 \end{bmatrix}.
\end{align}
We now use the assumption that the extrinsic calibration is constant to derive a useful equation relating $\bm{\Theta}$ to the relative motions $\mbf{V_{s_t}}$:
\begin{align}
\mbf{T}_{a_{t}} &= \mbf{T}_{b_{t}}\bm{\Theta},\\
\mbf{T}_{a_{t-1}}\mbf{V}_{a_t} &= \mbf{T}_{b_{t-1}}\mbf{V}_{b_t}\bm{\Theta},\\
\mbf{T}_{b_{t-1}}^{-1}\mbf{T}_{a_{t-1}}\mbf{V}_{a_t} &= \mbf{V}_{b_t}\bm{\Theta},\\
\bm{\Theta}\mbf{V}_{a_t} &= \mbf{V}_{b_t}\bm{\Theta}.
\end{align}
This final expression is found in many calibration papers~\cite{heller2014hand, condurache2016orthogonal} and will allow us to form a cost function based on the noisy relative motion measurements $\hat{\mbf{V}}_{a_t}$ and $\hat{\mbf{V}}_{b_t}$.

%\begin{equation}
%\hat{\mbf{T}}_{a_{t,t+1}}} =
% \begin{bmatrix}
% \hat{\mbf{R}\\
%-q^TB & q^Tq - \lambda,
%\end{bmatrix}
%\end{equation}

\subsection{QCQP Formulation}

We are now able to formulate a QCQP that seeks an optimal $\bm{\Theta}$ matrix: 
\begin{equation}\label{eq:primal}
	\normalfont
	\begin{aligned}
	   \underset{\mbf{R},\mbf{t}}{\text{minimize}}
	  & \quad J_{\mbf{R}} + J_{\mbf{t}}, \\
	   \text{subject to}
	  & \quad \mbf{R} \in \text{SO}(3),
	\end{aligned}
\end{equation}
where 
\begin{equation}\label{eq:cost_rot}
J_{\mbf{R}} = \sum_{i=1}^n \kappa_i\frob{\mbf{R}\mbf{R}_{a_i} - \mbf{R}_{b_i}\mbf{R}}^2
\end{equation}
is the rotation cost and 
\begin{equation}\label{eq:cost_trans}
J_{\mbf{t}} = \sum_{i=1}^n \tau_i\norm{\mbf{R}\mbf{t}_{a_i} + \mbf{t} - \mbf{R}_{b_i}\mbf{t} - \mbf{t}_{b_i}}{\scriptscriptstyle{2}}^2
\end{equation}
is the translation cost. The cost function is derived by expanding the expression $\frob{\bm{\Theta}\mbf{V}_{a_i} - \mbf{V}_{b_i}\bm{\Theta}}^2$ and weighting each term based on the rotational and translational measurement confidence parameters $\kappa_i$ and $\tau_i$. Since our approach is not an exact MLE derivation, these parameters must be estimated from measurement confidences returned by the egomotion estimation algorithm(s) employed via heuristics like covariance propagation rules through nonlinear functions~\cite{brookshire2012automatic}. In many cases this probabilistic information is unavailable and the parameters can be set to unity, corresponding to many geometric approaches in the literature~\cite{heller2014hand}.

Since the terms within both quadratic norms in the primal problem (\ref{eq:primal}) are linear or constant with respect to our optimization variables $\mbf{R}$ and $\mbf{t}$, the resulting cost function is quadratic. The constraint $\mbf{R} \in \text{SO}(3)$ can also be written as the following set of quadratic equations: 
\begin{align}
\mbf{R}^T\mbf{R} &= \mbf{I}, \\
\mbf{R}_i \times \mbf{R}_j &= \mbf{R}_k, \ i,j,k = \text{cyclic}(1,2,3),
\end{align}
where $\text{cyclic}(1,2,3)$ indicates the cyclic permutations (including identity) of the set $\{1,2,3\}$. These are orthogonality and ``right-handedness" constraints on $\mbf{R}$. While the handedness constraints distinguish $\text{SO}(3)$ from its complement in $\text{O}(3)$, $\text{O}(3)$ is the disjoint union of these sets and therefore the orthogonality constraints alone are often sufficient for estimation problems~\cite{rosen2016se}. However, we will see in Section \ref{sec:lagrange} that the handedness constraints are useful for our particular problem and solution method. 

We add an additional homogenizing variable $y$ that makes the objective and constraints purely quadratic (i.e., no linear or constant terms in the objective and no linear terms in the equality constraints). This trick leads to a simple derivation of the Lagrangian dual problem in Section \ref{sec:lagrange} as an SDP~\cite{carlone2015lagrangian, briales2017convex}. The homogenized primal problem has modified translational cost
\begin{equation}\label{eq:cost_trans_homog}
\tilde{J}_{\mbf{t}} = \sum_{i=1}^n \tau_i\norm{\mbf{R}\mbf{t}_{a_i} + \mbf{t} - \mbf{R}_{b_i}\mbf{t} - y\mbf{t}_{b_i}}{\scriptscriptstyle{2}}^2
\end{equation}
and includes the additional constraint $y^2 = 1$. Finally, the $\text{SO}(3)$ constraints become
\begin{align*}	
\mbf{R}^T\mbf{R} &= y^2\mbf{I},\\
\mbf{R}_i \times \mbf{R}_j &= y\mbf{R}_k, \ i,j,k = \text{cyclic}(1,2,3).
\end{align*}

%We stress that our cost function is not a MLE formulation...?

\section{Lagrangian Dual}\label{sec:lagrange}
This section closely follows the derivation in \cite{briales2017convex}, wherein the authors also seek to solve a QCQP for an optimization variable that is an element of $\text{SE}(3)$. We begin by simplifying the cost function of (\ref{eq:primal}) by reorganizing $\bm{\Theta}$ into a column vector
\begin{align}
\mbf{x} &= [\mbf{t}^T \ \mbf{r}^T \ y]^T, \\
\mbf{r} &= \text{vec}({\mbf{R}}),
\end{align}
where $\text{vec}(\cdot)$ ``unwraps" the columns of its matrix argument into a single column. This allows us to apply the identity~\cite{fackler2005notes}
\begin{equation}
\text{vec}(\mbf{A}\mbf{X}\mbf{B}) = (\mbf{B}^T \otimes \mbf{A}) \text{vec}(\mbf{X}) 
\end{equation}
to each term in the cost functions in Equations \ref{eq:cost_rot} and \ref{eq:cost_trans} to write the cost as a quadratic form,
\begin{equation}
J_\mbf{R} + \tilde{J}_\mbf{t} = \mbf{x}^T\mbf{Q}\mbf{x},
\end{equation}
where
\begin{equation}
\mbf{Q} = 
\begin{bmatrix}
   \mbf{0}_{3\times3} & \mbf{0}_{3\times10} \\
   \mbf{0}_{10\times3} & \mbf{Q}_\mbf{r}
\end{bmatrix}
+ \mbf{Q}_{\mbf{t}} \in \mathbb{R}^{13\times13},
\end{equation}
and
\begin{equation}
\mbf{Q}_\mbf{r} = 
\begin{bmatrix} 
   \sum_{i=1}^n \kappa_i\mbf{M}_{\mbf{r}, i}^T\mbf{M}_{\mbf{r}, i}  & \mbf{0}_{9\times 1} \\
   \mbf{0}_{1\times 9} & 0
\end{bmatrix},
\end{equation}
with each sub-matrix $\mbf{M}_{\mbf{r}, i}$ taking the form
\begin{equation}
\mbf{M}_{\mbf{r}, i} = (\mbf{R}_{a_i}^T \otimes \mbf{I}) - (\mbf{I} \otimes \mbf{R}_{b_i}) \in \mathbb{R}^{9\times9}.
\end{equation}
Similarly for the translation components:
\begin{align}
\mbf{Q}_\mbf{t} &= \sum_{i=1}^n \tau_i\mbf{M}_{\mbf{t}, i}^T\mbf{M}_{\mbf{t}, i}  \in \text{Sym}(13), \\
\mbf{M}_{\mbf{t}, i} &= [\mbf{I}-\mbf{R}_{b_i} \ \ (\mbf{t}_{a_i}^T \otimes \mbf{I}) \ \  {-\mbf{t}_{b_i}}].
\end{align}

Next, we note that the translation $\mbf{t}$ is unconstrained and can therefore be solved in closed form given the optimal rotation $\mbf{R}^\star$. Following the procedure in \cite{briales2017convex}, we can write 
\begin{equation}
\label{eq:bigQ}
\mbf{Q} = 
\begin{bmatrix}
\mbf{Q}_{\mbf{t}, \mbf{t}} & \mbf{Q}_{\mbf{t}, \tilde{\mbf{r}}} \\
\mbf{Q}_{\tilde{\mbf{r}}, \mbf{t}} & \mbf{Q}_{\tilde{\mbf{r}}, \tilde{\mbf{r}}}
\end{bmatrix},
\end{equation}
which allows us to concisely write the optimal translation
\begin{equation}
\label{eq:t_afo_R}
\mbf{t}^\star(\mbf{R}^\star) = -\mbf{Q}_{\mbf{t}, \mbf{t}}^{-1}\mbf{Q}_{\mbf{t}, \tilde{\mbf{r}}}\tilde{\mbf{r}}^\star,
\end{equation}
where $\tilde{\mbf{r}} = [\mbf{r}^T \ y]^T$. We can now substitute this expression for $\mbf{t}^\star$ into the cost function, allowing us to define 
\begin{equation} \label{eq:schur}
\tilde{\mbf{Q}} = \mbf{Q}/\mbf{Q}_{\mbf{t},\mbf{t}}.
\end{equation}
This leads to a simpler QCQP with only $\tilde{\mbf{r}}$ as a variable: 
\begin{equation}\label{eq:rotation_primal}
	\normalfont
	\begin{aligned}
	   \underset{\mbf{r}, y}{\text{minimize}}
	  & \quad \tilde{\mbf{r}}^T\tilde{\mbf{Q}}\tilde{\mbf{r}}, \\
	   \text{subject to}
	  & \quad \mbf{R} \in \text{SO}(3), \\
	  & \quad y^2 = 1.
	\end{aligned}
\end{equation}
 
\subsection{Dual Problem}
The Lagrangian dual problem of a homogeneous QCQP is a semidefinite program (SDP)~\cite{cifuentes2017local}. For details on deriving the Lagrangian dual of an optimization problem, see Chapter 5 of \cite{boyd2004convex}. Since the structure of (\ref{eq:rotation_primal}) is identical to the problem in \cite{briales2017convex} with the exception of a different $\tilde{\mbf{Q}}$ matrix, the dual problem is the following SDP which is derived using an identical  procedure:
\begin{equation}\label{eq:dual}
	\normalfont
	\begin{aligned}
	   \underset{\mathbf{M}, \bm{\lambda}, \gamma}{\text{minimize}}
	  & \quad \gamma, \\
	   \text{subject to}
	  & \quad  \tilde{\mbf{Q}} + \mbf{P}(\bm{\lambda}, \mathbf{M}, \gamma) \succeq \mbf{0}.
	\end{aligned}
\end{equation}
The only constraint is a linear matrix inequality (LMI) which is a function of the dual variables $\bm{\lambda} \in \mathbb{R}^{9 \times 1}$, $\mathbf{M} \in \text{Sym}(3)$, \vp{If $\mathbf{M}$ is a matrix, should it be capitalized?}  and $\gamma$ as well as the original problem data in the form of $\tilde{\mbf{Q}}$. The matrix $\mbf{P}(\bm{\lambda}, \mathbf{M}, \gamma)$ is a function of the constraints and corresponding dual variables on $\mbf{R}$ when written as a quadratic form like the cost function. The sum of $\tilde{\mbf{Q}}$ and  $\mbf{P}(\bm{\lambda}, \mathbf{M}, \gamma)$  forms the Hessian of the Lagrangian of the primal optimization problem. For details on the structure and derivation of $\mbf{P}(\bm{\lambda}, \mathbf{M}, \gamma)$, see \cite{briales2017convex} and its supplementary material~\cite{brialessupplementary}.
%\subsection{Strong Duality}

\subsection{Redundant Constraints}
Once again following the procedure in \cite{briales2017convex}, we note that while the column orthogonality constraints $\mbf{R}^T\mbf{R} = \mbf{I}$ and the row orthogonality constraints $\mbf{R}\mbf{R}^T = \mbf{I}$ are redundant descriptions for $\text{O}(3)$, they are linearly independent. In other words, the equations representing the column constraints cannot be written as a linear combination of the equations representing row constraints. In \cite{anstreicher2000lagrangian} and \cite{wolkowicz2002note}, the inclusion of both sets of independent orthogonality constraints is shown to produce a primal/dual problem pair with a smaller duality gap (without changing the primal problem's solution). To this end, we modify our primal by adding the redundant row orthogonality constraint. This leads to another dual variable $\mathbf{N} \in \text{Sym}(3)$ \vp{Same comment here} and a modified constraint matrix $\tilde{\mbf{P}}$ in our new ``strengthened" dual problem:
\begin{equation}\label{eq:dual_strengthened}
	\normalfont
	\begin{aligned}
	   \underset{\bm{\lambda}, \mathbf{M}, \mathbf{N}, \gamma}{\text{minimize}}
	  & \quad \gamma, \\
	   \text{subject to}
	  & \quad  \tilde{\mbf{Q}} + \tilde{\mbf{P}}(\bm{\lambda}, \mathbf{M}, \mathbf{N}, \gamma) \succeq \mbf{0}.
	\end{aligned}
\end{equation}
We summarize our approach in Algorithm \ref{alg:main_alg}, and visually outline the key steps in Figure \ref{fig:system}. 

\subsection{Observability and Strong Duality} \label{sec:observability}
The theorems in \cite{cifuentes2017local} provide methods of determining whether a particular QCQP with a known zero-duality-gap solution will exhibit strong duality when its problem parameters are perturbed (e.g., with noise in the case of sensor measurements). \vp{The next sentence needs to be reworded to the active 'We verify..'} Theorem 2 in our supplementary material was used in \cite{cifuentes2017local} to verify the empirical observation in \cite{carlone2015lagrangian} of strong duality for strictly convex noisy pose-graph optimization problems. Aside from strict convexity, this theorem requires that the noise-free problem's solution is an unconstrained optimizer, and that typical regularity conditions are satisfied. When these conditions are met, the existence of a zero-duality-gap region is theoretically guaranteed. It is easy to demonstrate numerically that most instances of (\ref{eq:rotation_primal}) with exact measurements (i.e., no noise) have zero-duality-gap. We present the following theorem outlining \textit{necessary} conditions for our cost function to exhibit strict convexity. If strict convexity is present, we demonstrate in the supplementary material that the other conditions required by Theorem 2 \jk{is it worth here giving a 1-2 sentence blurb about what Theorem 2 does - I get it after reading appendix, but it would be good to explain here or before exactly what Theorem 2 says... I may have just missed it...} can be demonstrated to hold via the methods in \cite{cifuentes2017local} for the example of SO$(d)$ synchronization. 

\begin{theorem} \label{thm:strong_duality}
An instance of our extrinsic calibration from egomotion problem has a strictly convex cost only if the measurement data contains rotations of the sensor rig about at least two unique axes. 
\end{theorem}

See the supplementary material for a proof of Theorem \ref{thm:strong_duality}. These necessary conditions are precisely the observability requirements derived in \cite{brookshire2013extrinsic} for the original formulation of the problem. Extensive numerical trials strongly suggest that the two-axis observability condition is also \textit{sufficient} for a strictly convex cost when combined with the requirement that both sensors are translated in the inertial world frame, however we leave the proof for future work. 

\begin{algorithm}
  \caption{Certifiably globally optimal calibration}
   \label{alg:main_alg}
   \begin{spacing}{1.1}
  \begin{algorithmic}[1]
    \Require{Relative motion measurements $\mathcal{V} = \{\hat{\mbf{V}}_{a_t}, \hat{\mbf{V}}_{b_t}\}_{t=1}^n$ }
    \Ensure{Global opt. extrinsic calibration $\bm{\Theta} = \{\mbf{R}^\star, \mbf{t}^\star\}$}
    \Function{Calibrate}{$\mathcal{V}$} 
   	\State Form $\mbf{Q}$ from $\mathcal{V}$  \Comment{\emph{Data matrix, Eq. (\ref{eq:bigQ})}}
    \Let{$\tilde{\mbf{Q}}$}{$\mbf{Q} / \mbf{Q}_{\mbf{t}, \mbf{t}}$} \Comment{\emph{Schur complement, Eq. (\ref{eq:schur})}}
    
    \State Form $\tilde{\mbf{P}}(\bm{\lambda}, \mathbf{M}, \mathbf{N}, \gamma)$ \Comment{Dual constraint matrix}
    \Let{$\mbf{R}^\star$}{Solution to SDP \Comment{\emph{Eq. (\ref{eq:dual_strengthened})}} }
    \Let{$\mbf{t}^\star$}{$-\mbf{Q}_{\mbf{t}, \mbf{t}}^{-1}\mbf{Q}_{\mbf{t}, \tilde{\mbf{r}}}\tilde{\mbf{r}}^\star$
}\Comment{\emph{Eq. (\ref{eq:t_afo_R})}, $\mbf{r}^\star = \text{vec}({\mbf{R}^\star})$}

    \State \Return{$\mbf{R}^\star, \mbf{t}^\star$}
    \EndFunction
  \end{algorithmic}
  \end{spacing}
\end{algorithm}

\section{Experiments}
\NewDocumentCommand\sigmaRot{}{\sigma_\mathrm{r}}
\NewDocumentCommand\sigmaTrans{}{\sigma_\mathrm{t}}
\NewDocumentCommand\Ltwo{}{L$^{\scriptstyle 2} \ $}
In this section we evaluate our algorithm's performance on synthetic and real data. We begin with an empirical analysis of the duality gap and the effect of adding redundant, independent constraints. Synthetic data is then used to compare the accuracy and runtime of our algorithm with the local optimization method in \cite{brookshire2013extrinsic}. Finally, we apply our algorithm to a real dataset provided by the authors of \cite{brookshire2013extrinsic} that uses RGB-D cameras to produce egomotion estimates. 

Throughout this section, we will frequently refer to the level of noise that is added to simulated measurements. We use $\sigmaTrans$ and $\sigmaRot$ to denote the standard deviation of translation in meters and rotation measurements in radians, respectively. To remain consistent with the experiments in \cite{brookshire2013extrinsic}, we apply zero-mean Gaussian noise to translations and an Euler angle representation of rotations. Since our cost function is not an MLE formulation and only supports scalar weights for translation and rotation costs, we use isotropic covariance matrices $\mbf{\Sigma}_\mathrm{t}=\sigmaTrans^2 \mbf{I}$ and $\mbf{\Sigma}_\mathrm{r} = \sigmaRot^2 \mbf{I}$. The performance metrics we use in this section are the \Ltwo norms for translation vectors (the Euclidean distance) and rotation matrices (the Frobenius norm). These metrics are used to compare the accuracy of estimates of $\bm{\Theta}$ acquired by our algorithm and the local optimization approach. 

\subsection{Simulated Data} \label{sec:simulated_data}

\begin{figure}[!h]
	\centering
	\includegraphics[width=0.48\textwidth]{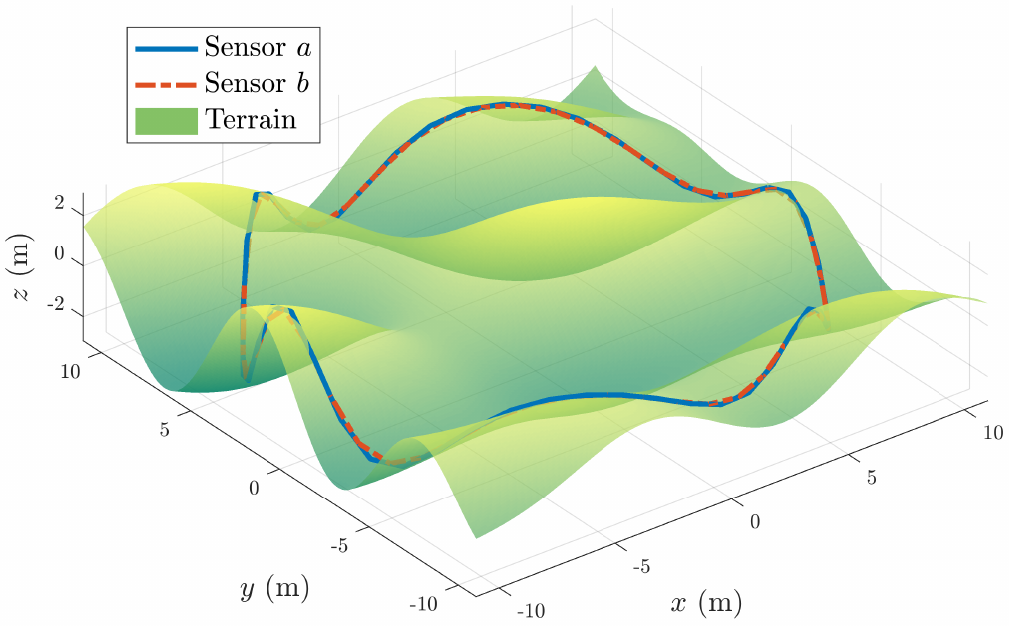}
	\caption{Simulated path example. A landscape generated with a mixture of random sinusoids is used to generate paths for sensors 1 and 2, which are rigidly fixed to a virtual vehicle traversing the terrain. A circular path over a terrain with varying elevation is chosen to ensure observability of the calibration parameters (see Section \ref{sec:observability}).}
	\label{fig:sim_path}
	\vspace{-0.2cm}
\end{figure}

To simulate measurements, we utilize a random smooth path that contains rotations about two distinct axes to guarantee that $\bm{\Theta}$ is observable~\cite{brookshire2013extrinsic}. An example path is displayed in Figure \ref{fig:sim_path}. The path is constructed by tracing a circular route over a landscape consisting of random sinusoidal functions in the z-axis in terms of x- and y-axis variables. The two sensor frames of reference are rigidly positioned according to their extrinsic calibration on a virtual vehicle which traverses this smooth path. The global poses of the individual sensors' paths are used to extract exact egomotion poses which are subsequently corrupted with noise. 

\subsubsection{Zero-Duality-Gap and Redundant Constraints}
\label{sec:duality_gap_redundant}
In Section \ref{sec:observability} we demonstrated that our problem formulation is guaranteed to  exhibit strong duality even in the presence of some finite measurement error (i.e., a zero-duality-gap region exists). In this section we empirically verify this result while also demonstrating that adding handedness and redundant orthogonality constraints increases the size of the zero-duality-gap region. While the exact amount of measurement error the problem can tolerate is difficult to determine for our problem's high-dimensional measurement data, a theorem in \cite{cifuentes2017local} provides a way of finding an analytical lower bound. We leave analysis of redundant constraints using these lower bounds as future work, and instead focus on experimental insights here. 

\begin{figure}[!h]
	\centering
	\includegraphics[width=0.48\textwidth]{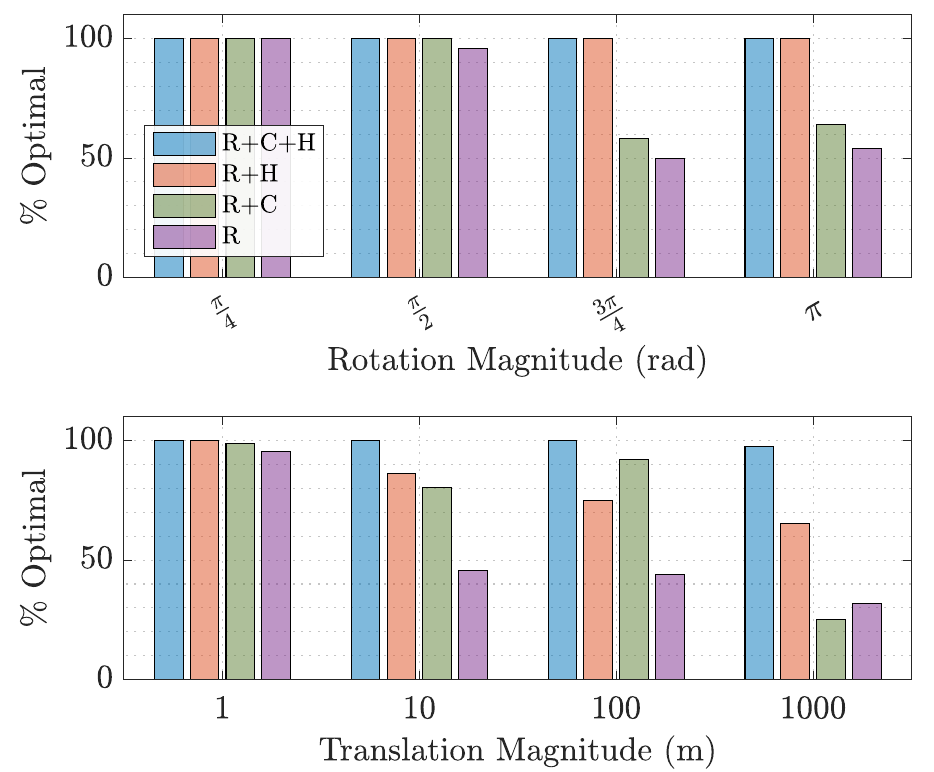}
	\caption{Performance of different O(3) and SO(3) constraints on a two-measurement simulated problem instance. All results used row orthogonality constraints at a minimum (R), with redundant column orthogonality (C) and right-handedness constraints (H) being added as well. In the top chart, each rotation magnitude (x-axis grouping) was used to create perturbations about 100 uniformly sampled axes of rotation. The bars indicate the percentage of certified globally optimal solutions found at a particular perturbation magnitude. The bottom chart has a fixed rotation perturbation magnitude of $\frac{\pi}{2}$ but also introduces a perturbation to one of the translation measurements over a total of 256 trials.}
	\label{fig:dualitygap}
	\vspace{0.1cm}
\end{figure}

In order to study the effect of adding constraints to the dual formulation, we formulated a simple observable problem instance: a rotation of $\frac{\pi}{2}$ radians and a translation of 1 meter about the vehicle's x-axis, followed by the same manoeuvre about the y-axis. Since our proof in the supplementary material guarantees a region of zero-duality-gap for even the minimally constrained case (i.e., only row orthogonality constraints), we expect adding constraints will improve the tolerance to measurement error. In order to keep the error injection process simple, we only perturbed one of the rotation measurements for each trial. The top bar graph in Figure \ref{fig:dualitygap} summarizes the effect of the magnitude of this rotation perturbation (i.e., the magnitude of the rotation angle in an axis-angle description of the perturbing rotation matrix). Each bar represents the percentage of 100 uniformly sampled axis directions that a given set of constraints returned for a fixed perturbation magnitude. For all noise levels tested, the handedness-augmented constraint sets (R+H and R+C+H) are able to recover an exact minimizer of the primal problem (i.e., strong duality holds), whereas the smaller constraint sets do not guarantee a zero-duality-gap problem instance. The redundant column orthogonality constraints (R+C) appear to improve the performance of the default O(3) constraints (R), but this benefit appears to be subsumed by the handedness constraints for all instances tested.

In the bottom bar graph of Figure \ref{fig:dualitygap}, the rotation perturbation magnitude was fixed to $\frac{\pi}{2}$ and a similar translation perturbation scheme was introduced for one of the measurements in our minimally observable problem. Since both rotation and translation directions needed to be sampled at a sufficient resolution, each bar represents 256 trials. Once severe translation error (10 m) is introduced, the benefit of including the complete, redundant constraint set (R+C+H) becomes clear. 

\subsubsection{Robustness to Noise}

\begin{figure}
    \centering
    \includegraphics[width=0.49\textwidth]{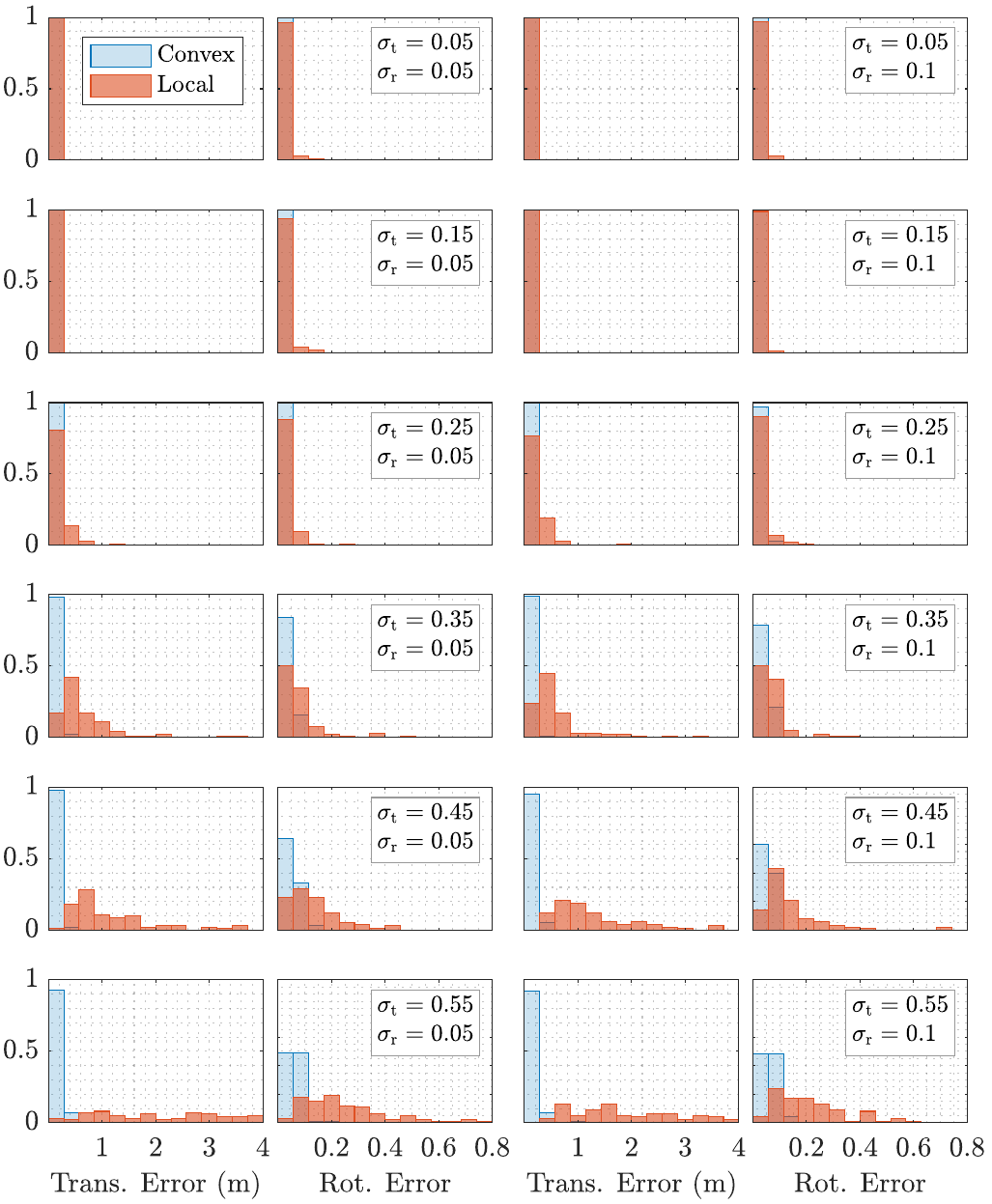}
    \caption{Histograms of translational and rotational \Ltwo error for various noise levels. The problem formulation and local solver from \cite{brookshire2013extrinsic} is shown in red, while our convex approach is in blue. Each subplot is a histogram showing the distribution of either translation or rotation error in the estimate of $\bm{\Theta}$ over 100 random trials. Rotation noise increases from left to right while translation noise increases from top to bottom. The performance of the local solver degrades much more rapidly as the noise levels increase. Note that rotation error is defined as the Frobenius norm of a rotation matrix and is therefore dimensionless.}
    \label{fig:tran_rot_hist}
    %\vspace{-0.2cm}
\end{figure}

Figure \ref{fig:tran_rot_hist} compares the performance of our method with the local optimization approach in \cite{brookshire2013extrinsic} for varying values of $\sigmaTrans$ and $\sigmaRot$. Each pair of subplots represents 100 random trials on simulated data. When $\sigmaTrans$ and $\sigmaRot$ are low, the calibration results are similar for both methods. Our approach is consistently more accurate than the local optimization approach, especially as the noise becomes extreme. This regime of noise is relevant for applications with low cost, noisy sensors or environments that produce challenging conditions (e.g. an urban canopy causing intermittent GPS readings or low-texture surfaces for camera-based egomotion estimation). 

We speculate that these results are due to local minima in the high-dimensional cost function of the approach in \cite{brookshire2013extrinsic}. Since that method uses relative pose measurements from one of the sensors as initial guesses for their corresponding variable in the optimization, noisy measurements mean that many of these initializations will be inaccurate. Our algorithm avoids this problem by treating all measurements as data instead of variables, and by solving and certifying the global optimality of a convex realization of the problem. 

%\paragraph{True Calibration}
\subsubsection{Initialization}
\begin{figure}[!h]
	\centering
	\includegraphics[width=0.48\textwidth]{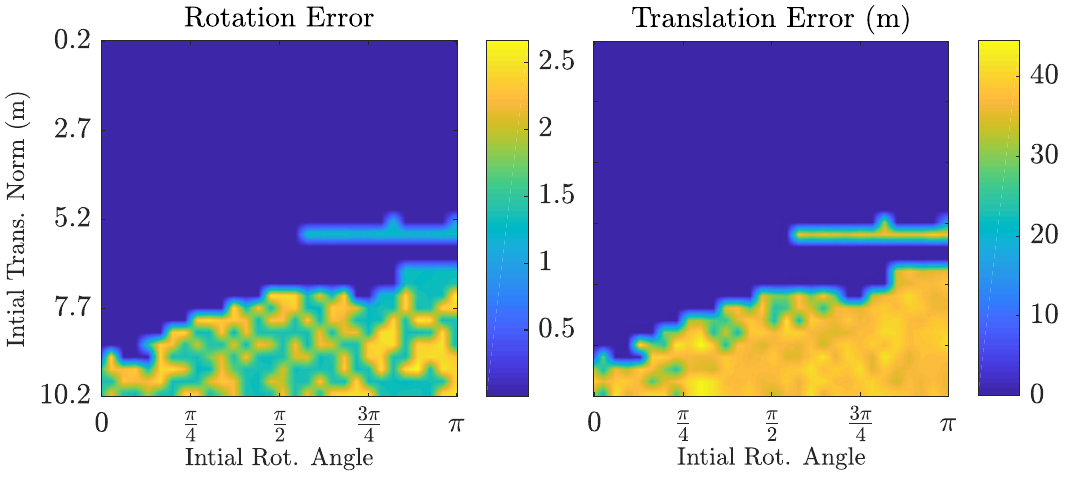}
	\caption{Heatmap displaying the difference in translational calibration error between the local approach and our convex algorithm for a randomly generated path. For each cell, 64 different initial values for $\bm{\Theta}$ were used to seed the local optimization. Each of these initial values had the same translation and and rotation angle magnitude, as indicated by the cell's position on the axes. The yellow colour plotted indicates higher error in the local optimization approach.}
	\label{fig:heatmap}
	
\end{figure}

For certain initial estimates of the extrinsic calibration $\bm{\Theta}$, the local optimization can also converge to a local minimum, even under mild noise conditions. In Figure \ref{fig:heatmap} we have plotted a heat map of the relative performance of the local approach and our global convex approach. Each cell corresponds to the maximum difference in the translation (left) or rotation (right) error of the two algorithms' estimates of $\bm{\Theta}$ over a uniform sampling of an initial guess for $\bm{\Theta}$. Higher values indicate larger error for the local approach. The x-axis varies the magnitude of the initial rotation guess' angle in an axis angle form, while the y-axis varies the magnitude of the initial translation guess. As the distance from the true calibration parameters increases, the accuracy of the local minima degrade. In contrast, the convex formulation converges for any initial conditions. 

\subsubsection{Runtime}
Since our problem formulation fixes the number of optimization variables, solver runtime is independent of the amount of data available. In contrast, the MLE approach in \cite{brookshire2013extrinsic} treats every relative motion as an optimization variable. We performed a runtime comparison of both methods on a desktop with an Intel Core I7-7820X 3.6 GHz CPU. Since both algorithms require data setup that is slow when naively implemented in MATLAB, we only compare the runtime of the solvers. Our approach uses the general-purpose SDPT3 algorithm~\cite{toh1999sdpt3} provided by the CVX package~\cite{grant2008cvx}. The local approach in \cite{brookshire2013extrinsic} uses a custom Levenberg-Marquardt solver. 

\begin{figure}[!h]
	\centering
	\includegraphics[width=0.49\textwidth]{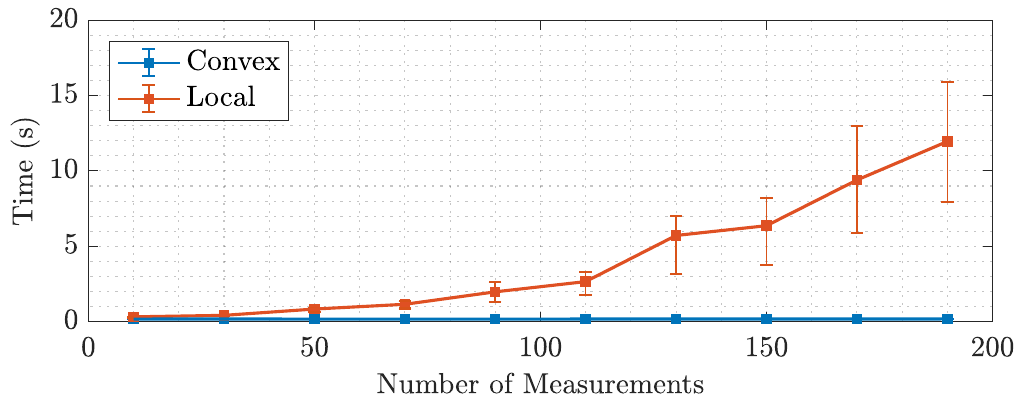}
	\caption{Mean solver runtime with 1st and 3rd quartile bars over 20 random runs per data point. The number of relative pose measurements in the problem data used is varied, exhibiting poor scaling for the local optimization approach and virtually constant solver time for our convex approach.}
	\label{fig:time_vs_N}
\end{figure}

In Figure \ref{fig:time_vs_N}, the mean runtime along with 1st and 3rd quartile bars from 100 random runs is plotted for both methods. The local optimization approach's runtime quickly grows as more measurements are added, while our approach is able to solve any problem instance in under half a second on average. The trend appears to continue for larger datasets, as the local approach did not converge after waiting several hours for a dataset of over 1000 poses, which our solver was able to easily handle. 

\subsection{RGB-D Data}

In this section, we use experimental data provided by the authors of \cite{brookshire2013extrinsic}, obtained with Xtion RGB-D sensors. For details about data collection, the extraction of relative pose measurements, and ground truth, see \cite{brookshire2013extrinsic}. In Table \ref{table:kinfu_err} we compare the error with respect to ground truth of our algorithm and the local optimization approach. Both algorithms are able to acquire accurate extrinsic calibrations. We also include the mean runtime of 100 applications of each solver on a laptop computer with an Intel Core i5-5287U 2.90 GHz CPU. For this dataset, we were able to find initial parameters that trap the local optimization method in local minima in a pattern similar to Figure \ref{fig:heatmap}, but only for initial parameters that are tens of meters away from the true value. Nevertheless, our method is much faster, and simulation results indicate that local minima arise for realistic datasets with different noise levels and types of measurements. Furthermore, an exhaustive set of experiments is impossible for the uncountably infinite number of problem instances, meaning our formal optimality guarantees are a valuable tool and safer choice.  \jk{Nonetheless, our method guarantees that... Why do we still win? ;-)}

\subsection{Starry Night Dataset}

\begin{figure}[t]
    \vspace{0.12cm}
    \centering
    \includegraphics[width=0.45\textwidth]{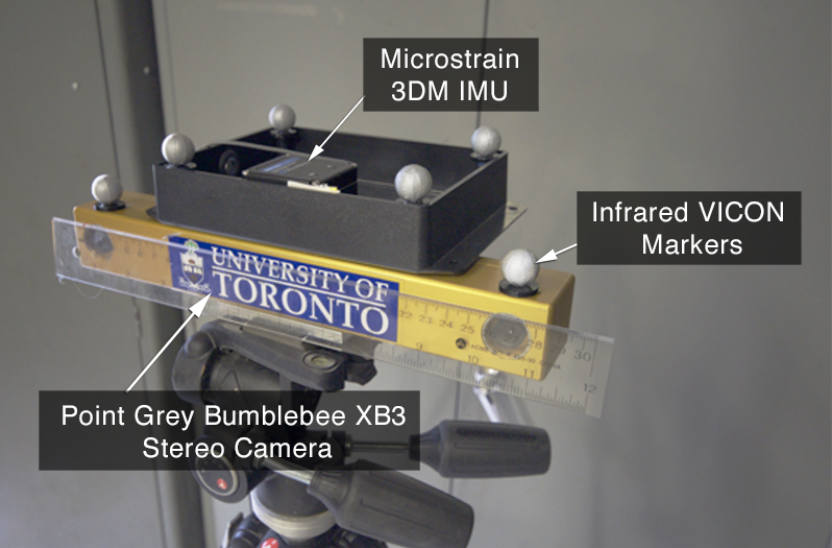}
    \caption{The sensor rig used in the ``Starry Night'' dataset. The IMU data reports translational and rotational velocities, while the stereo camera observes static point landmarks with known positions.}
    \label{fig:sensorhead}
\end{figure}

The ``Starry Night" dataset consists of stereo vision and pre-processed inertial measurement unit (IMU) readings from an environment with static landmarks~\cite{furgale2012continuous}. The dataset also contains Vicon motion capture measurements of the sensor rig's motion and a groundtruth value of $\mbf{\Theta}$. We used the dataset to produce incremental egomotion estimates for the stereo camera, IMU, and a second IMU trajectory estimate using Vicon measurements of reflective markers placed on the sensor rig. See Figure \ref{fig:sensorhead} for a photograph of the sensor rig.

Table \ref{table:kinfu_err} summarizes the results of a comparison between our algorithm and the local method of \cite{brookshire2013extrinsic}. The estimate of $\mbf{\Theta}$ that uses Vicon and stereo camera measurements is labelled ``Vicon", and both algorithms are able to obtain a fairly accurate estimate from this data. However, our algorithm is once again orders of magnitude faster. The results labelled ``IMU-5" compare integrated IMU measurements and camera localization on all poses where at least 5 out of 20 possible landmarks are visible to the stereo camera, ensuring accurate estimation. Even though the IMU measurements have been sanitized to remove biases, the velocity measurements are still very noisy and lead to large drift in the trajectory estimate. This is reflected in the results: both algorithms are able to recover a reasonable estimate of the extrinsic orientation, but neither manages to accurately estimate the relative position. The local approach is similarly far slower with this data. In order to study the accuracy of the algorithms on a number of poses where the local optimization approach has a competitive runtime, we downsampled the IMU and camera data to only those poses where the camera is able to see 15 landmarks. These results are labelled ``IMU-15", and neither algorithm is able to obtain an accurate estimate, but the local approach's result is particularly erroneous, which indicates that it may be trapped in a local minimum. Our globally optimal approach's rotational error is not egregiously large, and its fast runtime means it could be suitable for bootstrapping another algorithm that uses more data than relative poses but needs an initialization within some basin of convergence.
% Show effect of lowering number of poses? N_landmarks=5 at the moment, change to 

% TODO: add figure from Starry Night?

\begin{table}[h!]
	\centering
	\caption{Calibration Results On Real Data}
	\label{table:kinfu_err}
	\begin{threeparttable}
		
	\begin{tabular}{llccc}
	\toprule
		\textbf{Dataset} & \textbf{Method} & \multicolumn{2}{c}{\textbf{\Ltwo Error}} & \textbf{Runtime} (s)\\  
		
								& & Trans. (cm)    & Rot.    \\  \midrule%\cmidrule{3-4}
	\addlinespace[1.0ex]RGB-D	& Local & 1.2           & 0.026   & 4.80\\
		& Convex       & 1.3           & 0.026    & 0.31 \\ 
		
	\addlinespace[1.5ex]Vicon	& Local & 2.6          & 0.021   & 168.2\\ 
		& Convex       & 3.4           & 0.020    & 0.33 \\
	\addlinespace[1.5ex]IMU-5	& Local & 16.5           & 0.14   & 133.4 \\
		& Convex       & 21.3           & 0.17    & 0.40   \\
	\addlinespace[1.5ex]IMU-15	& Local & 709.7          & 2.30   & 0.90    \\
		& Convex       & 97.2           & 0.63    & 0.38    \\ \bottomrule
		\end{tabular}
		\end{threeparttable}
\end{table}

\section{Conclusion and Future Work}
\label{sec:conclusion}

In this work, we leveraged state-of-the-art certifiably globally optimal solution methods for QCQPs to create a novel, general purpose extrinsic calibration algorithm. Our method is faster than previous approaches, but more importantly it is able to avoid convergence to local minima that other methods fall prey to, even in the presence of severe noise. Our algorithm's beneficial properties were demonstrated through a range of experiments involving simulated and real data, which we intend to compare with a greater variety of algorithms in future work. We provided a proof that observability of the calibration parameters is a necessary condition for the guaranteed tightness of our dual solution, and we plan to prove it is also a sufficient condition. We also believe that the effect of redundant constraints can be more precisely characterized using the theory developed in \cite{cifuentes2017local}. 

The techniques used here can also be extended to the calibration of other sensor configurations. Monocular cameras are particularly interesting because they are only able to provide egomotion estimates up to an unknown scale. Additionally, the extrinsic calibration of multiple inertial measurement units could also benefit from the globally optimal properties of a QCQP formulation. We would also like to incorporate accurate, non-isotropic measurement covariances into a MLE formulation that can still be written as a QCQP, since this information, when available, can be extremely valuable to an estimator. Finally, a further avenue for future work is to perform a comparison of $\text{SO}(3)$ representations that admit a QCQP formulation of the problem (e.g., comparing rotation matrices with unit quaternions). 

\section*{Acknowledgement}
We would like to thank Prof. Timothy D. Barfoot for use of the
``Starry Night" dataset.

\bibliographystyle{IEEEtran} % use IEEEtran.bst style
\bibliography{calibration} % our bib file plus IEEE abreviation strings

\end{document}